\documentclass[letterpaper, 10 pt, journal, twoside]{IEEEtran} 

\IEEEoverridecommandlockouts                              

\usepackage{amsmath} 
\usepackage{amssymb}  
\usepackage{dsfont}
\usepackage{graphicx}

\usepackage{psfrag,graphicx,epsfig}
\usepackage{epstopdf}
\usepackage{xspace}
\usepackage{subfig}
\usepackage{float}
\usepackage{placeins}
\usepackage{multirow}
\usepackage{pgf,tikz}
\usepackage{nowidow}
\usepackage{lineno}
\usepackage{xcolor}

\usepackage{siunitx}
\usepackage{color}
\usepackage{flushend}
\usepackage{lineno}
\usepackage{algorithmic}
\usepackage[vlined,linesnumbered,boxruled]{algorithm2e}
\usepackage{enumitem}
\usepackage{caption}
\usepackage{tikz} 
\usepackage{tikz-3dplot}
\usetikzlibrary{fit,positioning}
\newcommand{\AxisRotator}[1][rotate=0]{%
	\tikz [x=0.25cm,y=0.60cm,line width=.2ex,-stealth,#1] \draw (0,0) arc (-150:150:1 and 1);%
}

\begin{document}

\title{\LARGE \bf Model-Based Reinforcement Learning for Physical Systems Without Velocity and Acceleration Measurements}

\author{Alberto Dalla Libera$^{*2}$, Diego Romeres$^{*1}$, Devesh K. Jha$^{1}$, Bill Yerazunis$^{1}$ and Daniel Nikovski$^{1}$
\thanks{$^{*}$These Authors contributed equally.}
	\thanks{$^{1}$Diego Romeres, Devesh K. Jha, Bill Yerazunis $\&$ Daniel~Nikovski are with Mitsubishi Electric Research Laboratories, Cambridge, MA 02139. Email--{\tt\small \{romeres,jha,yerazunis,nikovski\}@merl.com}}%
	\thanks{$^{2}$Alberto D. Libera is with Dept. of Information  Engineering, University of Padova, Via Gradenigo 6/b, 35131, Padova, Italy Email--{\tt\small dallaliber@dei.unipd.it}}%
}

\maketitle
 \thispagestyle{empty}
 \pagestyle{empty}

\begin{abstract}
\textcolor{black}{In this paper, we propose a derivative-free model learning framework for Reinforcement Learning (RL) algorithms based on Gaussian Process Regression (GPR). In many mechanical systems, only positions can be measured by the sensing instruments. Then, instead of representing the system state as suggested by the physics with a collection of positions, velocities, and accelerations, we define the state as the set of past position measurements. However, the equation of motions derived by physical first principles cannot be directly applied in this framework, being functions of velocities and accelerations. For this reason, we introduce a novel derivative-free physically-inspired kernel, which can be easily combined with nonparametric derivative-free Gaussian Process models. Tests performed on two real platforms show that the considered state definition combined with the proposed model improves estimation performance and data-efficiency w.r.t. traditional models based on GPR. Finally, we validate the proposed framework by solving two RL control problems for two real robotic systems.}
\end{abstract}


%
\IEEEpeerreviewmaketitle

\section{INTRODUCTION} \label{sec:intro}
\IEEEPARstart{R}{einforcement} Learning (RL) has seen explosive growth in recent years. RL algorithms have been able to reach and exceed human-level performance in several benchmark problems, such as playing the games of chess, go and shogi \cite{alphazero}. Despite these remarkable results, the application of RL to real physical systems (e.g., robotic systems) is still a challenge, because of the large amount of experience required and the safety risks associated with random exploration. 

To overcome these limitations, Model-Based RL (MBRL) techniques have been developed \cite{pilco, iLQG, GPS}. Providing an explicit or learned model of the physical system allows drastic decreases in the experience time required to converge to good solutions, while also reducing the risk of damage to the hardware during exploration and policy improvement.

Describing the evolution of physical systems is generally very challenging, and still an active area of research. Deriving models from first principles of physics might be very difficult, and could also introduce biases due to parameter uncertainties and unmodelled nonlinear effects. On the other hand, learning a model solely from data could be expensive, and generally suffers from insufficient generalization. Models based on Gaussian Process Regression (GPR) \cite{Rasmussen} have received considerable attention for model learning tasks in MBRL \cite{pilco}. GPR allows to merge prior physical information with data-driven knowledge, i.e., information inferred from analyzing the similarity between data, leading to so-called semi-parametric models \cite{romeres2016onlineIcub, romeres2019semiparametrical, ICRA2010NguyenTuong_62320}.

Physical laws suggest that the state of a mechanical system can be described by positions, velocities, and accelerations of its generalized coordinates. However, velocity and acceleration sensors are often not available, in particular when considering low-cost experimental setups. In such cases, velocities and accelerations are usually estimated by means of causal numerical differentiation of positions, introducing a difference between the real and estimated signals. These signal distortions can be seen as an additional unknown input noise, which might compromise significantly the prediction accuracy of the learning algorithm. Indeed, standard GPR models do not consider noisy inputs. Several Heteroscedastic GPR models have been proposed in the literature, see for example \cite{HGP_1,HGP_2,mchutchon2011gaussian}. However, the solutions proposed might not be suitable for real-time application, and most of the time they are more useful for improving the estimation of uncertainty, than for improving the accuracy of prediction. 

In this work, we propose a learning framework for model-based RL algorithms that does not need measurements of velocities and accelerations. Instead of representing the system state as a collection of positions, velocities, and accelerations, we propose to define the state as a finite past history of the position measurements. We call this representation derivative-free, to express the idea that the derivatives of position are not included in it.

The use of the past history of the state has been considered in the GP-NARX literature \cite{mchutchon2011gaussian,mattos2016latent,doerr2017optimizing}, \textcolor{black}{as well as in Eigensystem realization algorithm (ERA) and Dynamic Mode Decomposition (DMD) \cite{ERA,DMD}}. However, these techniques do not use a derivative-free approach when dealing with physical systems, e.g., they consider the history of position and velocity having double state dimension w.r.t. our approach (which might be a problem for MBRL) and do not incorporate prior physical model to design the covariance function. Derivative-free GPR models have also already been introduced in \cite{romeres2018DerivativeFree}, where the authors proposed derivative-free nonparametric kernels.

\textcolor{black}{The proposed approach has some connections with discrete dynamics models, see for instance \cite{DT1,DT2}. In these works, the authors derived a discrete-time model of the dynamics of a manipulator discretizing the Lagrangian equations. However, different from our approach, these techniques assume a complete knowledge of the dynamics parameters, typically identified in continuous time. Finally, such models might not be sufficiently flexible to capture unmodeled behaviors like delays,  backlash, and elasticity.}

\textbf{Contribution.} The main contribution of the present work is the formulation of derivative-free GPR models capable of encoding physical prior knowledge of mechanical systems that naturally depend
on velocity and acceleration. We propose physically inspired derivative-free (PIDF) kernels, which provide better generalization properties than the nonparametric deriviative-free kernel, and enable the design of semi-parametric derivative-free (SPDF) models.

{\color{black}The commonly used derivative and acceleration signals approximated through numerical differentiation represent statistics of the past raw positional data that cannot be exact, in general. The proposed framework does not make these computational assumptions, thus preserving richer information content in the inputs that are fed into the model. Moreover, providing to the GPR model a sufficient reach past history we can capture eventual higher orders unmodeled behaviors, like delays,  backlash, and elasticity.}

The proposed learning framework is tested on two real systems, a ball-and-beam platform and a Furuta pendulum. The experiments show that the proposed derivative-free learning framework improves significantly the estimation performance obtained by standard derivative-based models. \textcolor{black}{The SPDF models are used to solve RL-based trajectory optimization tasks. In both systems, we applied the control trajectory obtained by an iLQG~\cite{tassa2012synthesis} algorithm in an open-loop fashion. The obtained performance shows that the proposed framework learns accurately the dynamics of the two systems, and  it is suitable for RL applications.}

The paper is organized as follows. In Section \ref{sec:GPR}, we briefly introduce the standard learning framework adopted in model-based RL using GPR. Then, in Section \ref{sec:proposed_apporach}, we propose our derivative-free learning framework composed of the definition of the state and a novel derivative-free prior for GPR, based on the physical equations of motion. Finally, in the last two sections, we report the performed experiments.

\section{MODEL BASED REINFORCEMENT LEARNING USING GAUSSIAN PROCESS REGRESSION} \label{sec:GPR}
In this section, we describe the standard model learning framework adopted in MBRL using GPR, and the trajectory optimization algorithm applied. An environment for RL is formally defined by a Markov Decision Process (MDP). Consider a discrete-time system $\boldsymbol{\tilde{x}}_{k+1} = \boldsymbol{f}\left(\boldsymbol{\tilde{x}}_k, \boldsymbol{u}_k\right)$ subject to the Markov property, where $\boldsymbol{\tilde{x}}_{k} \in \mathbb{R}^{n_s}$ and $\boldsymbol{u}_{k} \in \mathbb{R}^{n_u}$ are the state vector and the input vector at the time instant $k$.

When considering a mechanical system with generalized coordinates $\boldsymbol{q}_k =  \left[q_k^1,\dots,q_k^n\right]\in \mathbb{R}^n$, the dynamics equations obtained through Rigid Body Dynamics, see \cite{siciliano2010robotics}, suggest that, in order to satisfy the Markov property, the state vector $\boldsymbol{\tilde{x}}$ should consist of positions, velocities, and accelerations of the generalized coordinates, i.e., $\boldsymbol{\tilde{x}}_k=\left[\boldsymbol{q}_k,\,\boldsymbol{\dot{q}}_k,\,\boldsymbol{\ddot{q}}_k  \right] \in \mathbb{R}^{3n}$, or possibly of a subset of these variables, depending on the task.

Model-based RL algorithms derive the policy $\pi\left(\boldsymbol{\tilde{x}}_k\right)$ starting from $\boldsymbol{\hat{f}}\left(\boldsymbol{\tilde{x}}_k, \boldsymbol{u}_k\right)$, an estimate of the system evolution. 

\subsection{Gaussian Process Regression}
\label{subsec:GPR}
In some studies, GPR \cite{Rasmussen} has been used to learn $\boldsymbol{\hat{f}}\left(\boldsymbol{\tilde{x}}_k, \boldsymbol{u}_k\right)$, see for instance \cite{pilco}. Typically, the variables composing $\boldsymbol{\tilde{x}}_{k+1}$ are assumed to be  conditionally independent given $\boldsymbol{\tilde{x}}_k$ and $\boldsymbol{u}_k$, and each state dimension is modeled by a separate GPR. The components of $\boldsymbol{\hat{f}}\left(\boldsymbol{\tilde{x}}_k, \boldsymbol{u}_k\right)$, denoted by $\hat{f}^i\left(\boldsymbol{\tilde{x}}_k, \boldsymbol{u}_k\right)$, with $i=1 \dots n_s$, are inferred and updated based on $\left\{X, \boldsymbol{y}^i\right\}$, a data set of input-output noisy observations. 
Let $N$ be the number of samples available, and define the set of GPR inputs as $X=\left[ \boldsymbol{\bar{x}}_1,\, \dots,\, \boldsymbol{\bar{x}}_N \right]$ where $\boldsymbol{\bar{x}}_k = \left[\boldsymbol{\tilde{x}}_k,\boldsymbol{u}_k\right] \in \mathbb{R}^{m}$ with $m=n_s+n_u$. As regards the outputs $\boldsymbol{y}^i = \left[y^i_{1},\, \dots,\, y^i_{N}\right]$, two definitions have been proposed in the literature. In particular, $y^i_{k}$ can be defined as $\tilde{x}^i_{k+1}$, the $\emph{i-th}$ component of the state at the next time instant, or as $y^i_{k}=\tilde{x}^i_{k+1}-\tilde{x}^i_{k}$, leading to $\boldsymbol{\hat{\tilde{x}}}_{k+1} = \boldsymbol{\tilde{x}}_{k} + \boldsymbol{\hat{f}}\left(\boldsymbol{\tilde{x}}_{k}, \boldsymbol{u}_{k}\right)$. In both cases, GPR models the observations as 
\begin{align}
\boldsymbol{y}^i &= \begin{bmatrix}
f^i(\boldsymbol{\bar{x}}_1), \ldots , f^i(\boldsymbol{\bar{x}}_N)
\end{bmatrix}^\top + \begin{bmatrix}
e_1 , \ldots , e_N
\end{bmatrix}^\top \notag \\ &= \boldsymbol{f}^i(X) + \boldsymbol{e} \text{,}
\label{eq:model_general}
\end{align}
where $\boldsymbol{e}$ is Gaussian i.i.d. noise with zero mean and covariance $\sigma^2_n$, and $\boldsymbol{f}^i(X) \sim \mathcal{N}\left( \boldsymbol{m_{f^i}}(X), K_{f^i}\left(X,X\right)\right)$. The matrix $K_{f^i}(X,X) \in \mathbb{R}^{N \times N}$ is called the kernel matrix, and is defined through the kernel function $k_{f^i}(\cdot,\cdot)$, i.e., the $K(X,X)$ entry in position $k$,$j$ is equal to $k(\boldsymbol{\bar{x}}_k,\boldsymbol{\bar{x}}_j)$. 
In GPR, the crucial aspect is the selection of the prior functions for $f^i(\cdot)$, defined by $m_{f^i}(\cdot)$, usually considered $0$, and $k_{f^i}(\cdot,\cdot)$. Then, see \cite{Rasmussen}, the maximum a posteriori estimator is:
\begin{equation}
    \hat{f}^i(\cdot) = K_{f^i}(\cdot,X)\left(K_{f^i}\left(X,X\right) + \sigma^2_n I_N \right)^{-1}\boldsymbol{y}^i,\label{eq:posterior_mean}
\end{equation}
In the following, we will refer to $f\left(\cdot\right)$ and $k\left(\cdot,\cdot\right)$ as one of the $\boldsymbol{f}(\cdot)$ components and the relative kernel function. 

\textbf{Physically inspired kernels}. When the physical model of the system is available, the model information might be used to identify a feature space over which the evolution of the system is linear. 
More precisely, assume that the model can be written in the form $y_k = \boldsymbol{\phi}(\boldsymbol{\bar{x}}_k)^T\boldsymbol{w}$, where $\boldsymbol{\phi}(\boldsymbol{\bar{x}}_k):\mathbb{R}^m \to \mathbb{R}^q$ is a known nonlinear function that maps the GPR inputs vector $\boldsymbol{\bar{x}}_k$ onto the physically inspired features space, and $\boldsymbol{w}$ is the vector of unknown parameters, modeled as a zero mean Gaussian random variable, i.e., $\boldsymbol{w} \sim N\left(0, \Sigma_{PI}\right)$, with $\Sigma_{PI} \in \mathbb{R}^{q \times q}$.
The expression of the physically inspired kernel (PI) is 
\begin{equation}
\label{eq:kernel_PI_general}
k(\boldsymbol{\bar{x}}_k,\boldsymbol{\bar{x}}_j) = \boldsymbol{\phi}(\boldsymbol{\bar{x}}_k)^T\Sigma_{PI}\boldsymbol{\phi}(\boldsymbol{\bar{x}}_j) \text{,}
\end{equation}
namely, a linear kernel in the features $\boldsymbol{\phi}(\cdot)$. For later convenience, we define also the homogeneous polynomial kernel in $\boldsymbol{\phi}(\cdot)$, which is a more general case of \eqref{eq:kernel_PI_general}: $k^p_{poly}(\boldsymbol{\bar{x}}_k,\boldsymbol{\bar{x}}_j) = \left( \boldsymbol{\phi}(\boldsymbol{\bar{x}}_k)^T\Sigma_{PI}\boldsymbol{\phi}(\boldsymbol{\bar{x}}_j) \right)^p \text{.}$

\textbf{Nonparametric kernel}. When a physical model is not available, the kernel has to be chosen by the user according to their understanding of the process to be modeled \cite{Rasmussen}. A common option is the Radial Basis Function kernel (RBF):
\begin{equation}
\label{eq:RBF_kernel}
k_{RBF}(\boldsymbol{\bar{x}}_k,\boldsymbol{\bar{x}}_j) = \lambda exp\left(-0.5||\boldsymbol{\bar{x}}_k-\boldsymbol{\bar{x}}_j||^2_{\Sigma_{RBF}}\right) \text{,}
\end{equation}
where $\lambda$ is a positive constant called the scaling factor, and $\Sigma_{RBF}$ is a positive definite matrix that defines the norm over which the distance between $\boldsymbol{\bar{x}}_k$ and $\boldsymbol{\bar{x}}_j$ is computed, i.e., $||\boldsymbol{x}||^2_{\Sigma_{RBF}}=\boldsymbol{x}^T\Sigma_{RBF}\boldsymbol{x}$. 
Several options to parameterize $\Sigma_{RBF}$ have been proposed, e.g., a diagonal matrix or a full matrix defined by the Cholesky decomposition, namely, $\Sigma_{RBF}=LL^T$\textcolor{black}{,  see \cite[Chp.5]{Rasmussen},\cite[Sec. 4.1]{6796297}}.

\textbf{Semiparametric kernel}. This approach combines the physically inspired and the non-parametric kernels. Here we define the kernel function as the sum of the covariances:
\begin{equation}
\label{eq: SP_kernel_general}
k(\boldsymbol{\bar{x}}_k, \boldsymbol{\bar{x}}_j) = \phi(\boldsymbol{\bar{x}}_k)^T\Sigma_{PI}\phi(\boldsymbol{\bar{x}}_j) + k_{NP}(\boldsymbol{\bar{x}}_k, \boldsymbol{\bar{x}}_j) \text{,}
\end{equation}
where $k_{NP}(\cdot, \cdot)$ can be, for example, the RBF kernel \eqref{eq:RBF_kernel}.

\vspace{-1mm}
\subsection{Trajectory Optimization using iLQG}
\label{subsec:iLQG}

The iLQG algorithm is a popular technique for trajectory optimization \cite{tassa2012synthesis}. Given discrete time
dynamics such as \eqref{eq:model_general} and a cost function, the algorithm computes locally linear models and quadratic cost functions for the system along a trajectory. These linear models are then used to compute optimal control inputs and local gain matrices by iteratively solving the associated LQG problem, 
 see~\cite{tassa2012synthesis}.
\section{DERIVATIVE-FREE FRAMEWORK FOR REINFORCEMENT LEARNING ALGORITHMS} \label{sec:proposed_apporach}

A novel learning framework to model the evolution of a physical system is proposed, which addresses several limitations of the standard modelling approach described in Sec.~\ref{sec:GPR}.\\
\textbf{Numerical differentiation.} The Rigid Body Dynamics of any physical system are functions of joint positions, velocities, and accelerations. However, a common issue is that often joint velocities and accelerations cannot be measured. 
\textcolor{black}{Computing them by means of causal numerical differentiation starting from the (possibly noisy) measurements of the joint positions might introduce considerable delays and distortions of the estimated signals. This fact could severely hamper the final solution.	This is a very well known and often discussed problem, see, e.g., \cite{siciliano2010robotics,hollerbach2008model,Nguyen-Tuong2011}.}\\
\textbf{Conditional Independence.} The assumption of conditional independence among the $f^i\left(\boldsymbol{\bar{x}}_k\right)$ with $i=1 \dots d$ given $\boldsymbol{\bar{x}}_k$ in \eqref{eq:model_general} might be a very imprecise approximation of the real system's behavior, in particular when the outputs considered are position, velocity, or acceleration of the same variable, which are correlated by nature. This fact has been shown to be an issue in estimation performance in \cite{romeres2019semiparametrical}, where the authors proposed to learn the acceleration function and integrate it forward in time in order to estimate position and velocity. 
Moreover, under this assumption, a separate GP for each output needs to be estimated for modeling variables that are intrinsically correlated, leading to redundant modeling design and testing work, and a waste of computational resources and time. This last aspect might be particularly relevant when considering systems with a considerable number of DoF.\\
\textbf{Delays and nonlinearities.}  Finally, physical systems are often affected by intrinsic delays and nonlinear effects that have an impact on the system over several time instants, contradicting the first-order Markov assumption; an instance of such behavior is reported in section \ref{sec:FP_models}.

\subsection{Derivative-Free State definition}
To overcome the aforementioned limitations, we define the system state\footnote{The exact state of a physical system is usually unknown, but in general accepted to be given by position, velocity and acceleration accordingly to the physics first principles. With a slight abuse of notation, we refer to our representation of the state in a derivative-free fashion as the state variable.} in a derivative-free fashion, considering as state elements the history of the position measurements:
\begin{equation}\label{eq:DF_state}
\boldsymbol{x}_k := \left[\boldsymbol{q}_k,\,\dots,\,\boldsymbol{q}_{k-k_p}\right] \in \mathbb{R}^{n(k_p+1)} \text{ , with } k_p \in \mathbb{Z}^+ \text{ .}
\end{equation}

The simple yet exact idea behind this definition is that when velocities and accelerations measures are not available, if $k_p$ is chosen sufficiently large, then the history of the positions contains all the system information available at time $k$, leaving to the model-learning algorithm the possibility of estimating the state transition function. Indeed, velocities and accelerations computed through causal numerical differentiation are the outputs of digital filters with finite impulse response (or with finite past instants knowledge for non-linear filters), which represent a statistic of the past raw position data. 
Notice that these statistics cannot be exact in general, leading to a loss of information that instead is kept in the proposed derivative-free framework.

The state transition function becomes deterministic and known (i.e., the identity function) for all the $[\boldsymbol{q}_{k-1},\,\dots,\,\boldsymbol{q}_{k-k_p}]$ components of the state. 
Consequently, the problem of learning the evolution of the system is restricted to learning only the functions $\boldsymbol{q}_{k+1}=\boldsymbol{f}\left(\boldsymbol{x}_k,\boldsymbol{u}_k\right)$, reducing the number of models to learn and avoiding erroneous conditional independence assumptions. Finally, the MDP has a state information rich enough to be robust to intrinsic delays and to obey the first-order Markov property.

\subsection{State Transition Learning with PIDF Kernel} \label{sec:guidelines}
Derivative-free GPRs have already been introduced in \cite{romeres2018DerivativeFree}, where the authors derived a data-driven derivative-free GPR. As pointed out in the introduction, the generalization performance of data-driven models might not be sufficient to guarantee robust learning performance, and exploiting eventual prior information coming from the physical model is crucial. 
To address this problem, we propose a novel Physically Inspired Derivative-Free (PIDF) kernel. 

The PIDF exploits the property that the product and sum of kernels is still a kernel, see \cite{Rasmussen}. Define $\boldsymbol{q}^i_{k^-}=\left[q_k^i,\dots,q_{k-k_p}^i\right] $
and assume~that a physical model of the type  $y_k =\boldsymbol{\phi}\left(\boldsymbol{q}_k, \boldsymbol{\dot{q}}_k, \boldsymbol{\ddot{q}}_k, \boldsymbol{u}_k \right)\boldsymbol{w}$, is known. Then, we propose a set of guidelines  to derive a PIDF kernel starting from
$\boldsymbol{\phi}$:

    \textbf{PIDF Kernel Guidelines}
       \begin{enumerate}[leftmargin=1.2em]
	\item Each and every position, velocity, or acceleration term in $\boldsymbol{\phi} (\cdot)$ is replaced by a distinct polynomial kernel $k_{poly}^p(\cdot,\cdot)$ of degree $p$, where $p$ is the degree of the original term; e.g., $\ddot{q}^{i^2} \rightarrow k_{poly}^2(\cdot,\cdot)$.
	\item The input of each of the kernels $k_{poly}^p(\cdot,\cdot)$ in 1) is a function of $\boldsymbol{q}^i_{k^-}$, the history of the position $q^i$ corresponding to the independent variable of the substituted term; \\ 
	e.g., $\ddot{q}^{i^2} \rightarrow k_{poly}^2(q^i_{k^-},\cdot)$.
	\item If a state variable appears into $\boldsymbol{\phi} (\cdot)$ transformed by a function $g(\cdot)$, the input to $k_{poly}^p(\cdot,\cdot)$ becomes the input defined at point 2) transformed by the same function $g(\cdot)$, e.g.,  $\sin(q^i) \rightarrow k_{poly}^1(\sin(\boldsymbol{q}^i_{k^-}),\sin(\boldsymbol{q}^i_{j^-}\cdot))$.
\end{enumerate}%
Applying these guidelines will generate a kernel function $k_{PIDF}(\cdot,\cdot)$, which incorporates the information given by the physics, without knowing the velocity and acceleration.

The extension to semiparametric derivative-free (SPDF) kernels becomes trivial. Combining, as described in Section~\ref{subsec:GPR}, the proposed $k_{PIDF}(\cdot,\cdot)$ with a derivative-free NP kernel, $k_{NPDF}(\cdot,\cdot)$ (or as proposed in \cite{romeres2018DerivativeFree}), we obtain:
\begin{equation}
\label{eq:SP_kernel_DF}
k_{SPDF}(\cdot, \cdot) = k_{PIDF}(\cdot, \cdot) + k_{NPDF}(\cdot, \cdot) \text{.}
\end{equation}

These guidelines formalize the solution to the non-trivial issue of modeling real systems using
physical models without measuring velocity and acceleration. \textcolor{black}{Although the guidelines might not be the only possible solution, they represent an algorithm with no ambiguity or arbiter choice to be made by the user to convert RBD into derivative free models.}

In the next sections, we apply the proposed learning framework to the benchmark systems Ball-and-Beam (BB) and Furuta Pendulum (FP), describing in detail the kernel derivations. While for both setups we will show the task of controlling the system, highlighting the advantages of the proposed derivative-free framework, due to space limitations, we decided to present different properties of the proposed method in each of them. In the BB case, we will highlight the estimation performance of $k_{PIDF}(\boldsymbol{x}_k,\cdot)$ over $k_{PI}(\boldsymbol{\tilde{x}}_k,\cdot)$ computing $\boldsymbol{\tilde{x}}_k$ with several filters and the difficulty of choosing the most suitable velocity. In the~more complex FP system, we analyze  robustness to delays, performance at $k-$step-ahead prediction, and make extensive comparisons among physically inspired, nonparametric, semiparametric derivative-free, and standard GPR. 

\section{BALL-AND-BEAM PLATFORM}
\begin{figure}[b]
	\vspace{-2mm}
	\centering
	\includegraphics[width=0.8\linewidth]{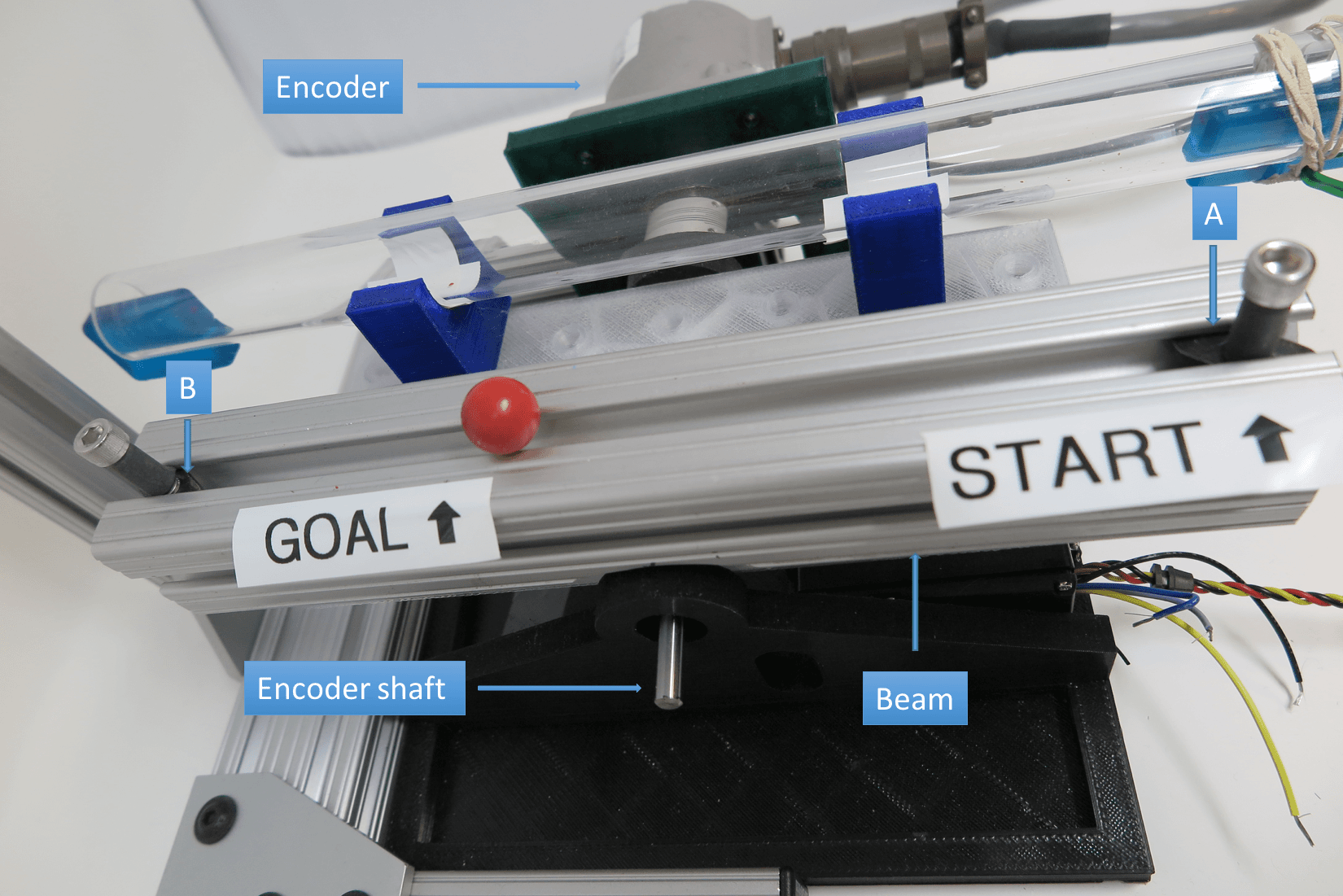}
	\caption{In-house built Ball-and-Beam experimental setup.}\label{fig:BB_setup}
	\vspace{-2mm}
\end{figure}
	Fig.~\ref{fig:BB_setup} shows our experimental setup for the BB system~\cite{8285376}. An aluminum bar is attached to a tip-tilt table (platform) constrained to have 1 degree of freedom (DoF). The platform is actuated by an inexpensive, commercial off-the-shelf HiTec type HS-805BB RC model servo motor that provides open-loop positioning; the platform angle is measured by an accurate absolute encoder. There is no tachometer attached to the axis, so angular velocity is not directly measurable. 
	A ball is rolling freely in the groove. We use an RGB camera which is attached to a fixed frame to measure the ball's position. The ball is tracked in real-time using a simple, yet fast, blob tracking algorithm. All the communication with the camera and servo motors driving the system is done using ROS \cite{quigley2009ros}.
	
	Let $\theta$ and $p$ be the beam angle and the ball position, respectively, considered in a reference frame with origin at the beam center and oriented s.t. the $A$ beam end is positive. The forward dynamics of the ball are expressed by the following equation (see \cite{BBmodel} for the details)
    \begin{align}
	\ddot{p} &= \left( m(p-l/2)\dot{\theta}^2 - mg\sin(\theta) - b\dot{p}\right)/\left(J_b/r^2 + m\right) \label{eq:BB_physical_model}
	\\ & = \left[p\dot{\theta}^2, \dot{\theta}^2,\,\sin(\theta),\, \dot{p} \right] \boldsymbol{w}  \notag = \boldsymbol{\phi}(\theta, \dot{\theta},p,\dot{p})^T \boldsymbol{w} \text{,}
	\end{align}  
	where $m$, $J_b$, $r$ and $b$ are the ball mass, inertia, radius, and friction coefficient, respectively. Starting from eq. \eqref{eq:BB_physical_model}, the forward function for $\Delta_{p_k} = p_{k+1}-p_k$ is derived by integrating $\ddot{p}$ twice forward in time, and assuming a constant $\ddot{p}$ between two time instants:
	\begin{equation}
	\Delta_{p_{k+1}} = \dot{p}_k\delta_t + \frac{\delta_t^2}{2}\boldsymbol{\phi}(\theta, \dot{\theta}, p, \dot{p})^T\boldsymbol{w} =\boldsymbol{\phi}(\theta, \dot{\theta}, p, \dot{p})^T\boldsymbol{w'} \text{,}\label{eq:linear_relation_delta_BB}
	\end{equation}
	where $\delta_t$ is the sampling time. In order to describe the BB system in the framework proposed in Section \ref{sec:proposed_apporach}, we define the derivative-free state as $\boldsymbol{x}_k = \left[\boldsymbol{x^p}_k,\,\boldsymbol{x^{\theta}}_k\right]$, with
	\begin{equation*}
		\boldsymbol{x^p}_k = \left[p_k,\,\dots,\,p_{k-k_p}\right] \text{,\,\,\,\,} 
        \boldsymbol{x^{\theta}}_k = \left[\theta_k,\,\dots,\,\theta_{k-k_p}\right] \text{.}
	\end{equation*}
	Applying the guidelines defined in section \ref{sec:guidelines} to Eq.~\eqref{eq:linear_relation_delta_BB}, the PIDF kernel obtained is
	\begin{align}
	&k^{BB}_{PIDF}\left(\boldsymbol{x}_k, \boldsymbol{x}_j\right) = k_{poly}^1\left(\boldsymbol{x^{p}}_k, \boldsymbol{x^{p}}_j\right) k_{poly}^2\left(\boldsymbol{x^{\theta}}_k, \boldsymbol{x^{\theta}}_j\right) + \nonumber \\
	&\quad k_{poly}^2\left(\boldsymbol{x^{\theta}}_k, \boldsymbol{x^{\theta}}_j\right) + k_{poly}\left(\sin\left(\boldsymbol{x^{\theta}}_k\right), \sin\left(\boldsymbol{x^{\theta}}_j\right)\right) + \nonumber \\
	&\quad + k_{poly}^1\left(\boldsymbol{x^{p}}_k, \boldsymbol{x^{p}}_j\right) \text{.} \label{eq:PIDF_BB}
	\end{align}
\subsection{Prediction performance}
\label{subsec:BB_pred_performance}
The purpose of this section is to compare the prediction performance of the GP models \eqref{eq:posterior_mean}, using as prior the PIDF kernel \eqref{eq:PIDF_BB}, $f_{PI_{DF}}(\boldsymbol{x})$, and using the standard PI kernel applying \eqref{eq:BB_physical_model} to Eq.~\eqref{eq:kernel_PI_general}, $f_{PI}(\boldsymbol{\tilde{x}})$. The question that the standard approach imposes is how to compute the velocities from the measurements in order to estimate $\boldsymbol{\tilde{x}}$, and there is not a unique answer to this question. We experimented with some common filters using different gains in order to find good velocity approximations: 
	\begin{itemize}[leftmargin=1.em]
		\item Standard numerical differentiation followed by a low pass filter to reject noise, which uses the position history $k_p = 5$. We considered 3 different cutoff frequencies $15$, $21$, $28$ Hz with correspondent estimators denominated as $f_{PI_{n1}}$, $f_{PI_{n2}}$, $f_{PI_{n3}}$, respectively;
		\item Kalman filter, with different process covariances $\Sigma_{x}=diag\left(\left[\sigma_x,\,\,2\sigma_x\right]\right)$ and $\sigma_x$ equals to $0.005$, $0.5$, $10$ with correspondent estimators $f_{PI_{KF1}}$, $f_{PI_{KF2}}$, $f_{PI_{KF3}}$;
		\item The acausal Savitzky-Golay filter $f_{PI_{SG}}$ with window length $5$.
	\end{itemize}
    Acausal filters have been introduced just to provide an upper bound on prediction performance; otherwise, they can not be applied in real-time applications. As regards the number of past time instants considered in $f_{PI_{DF}}$, we set $k_p=4$. 
    \textcolor{black}{Both the training and test datasets consists in the collection of 3 minutes of operation on the BB system, with control actions applied at 30Hz, while measurements from the encoder and camera were recorded. Both the datasets account for $5400$ samples. The control actions were generated as a sum of 10 sine waves with randomly sampled frequency between $[0,10]$ Hz, shift phases in $[0,2\pi]$, and amplitude ranging $\pm 5 [deg]$}.
    \begin{figure}[t]
    	\centering
    	\includegraphics[width=.95\columnwidth]{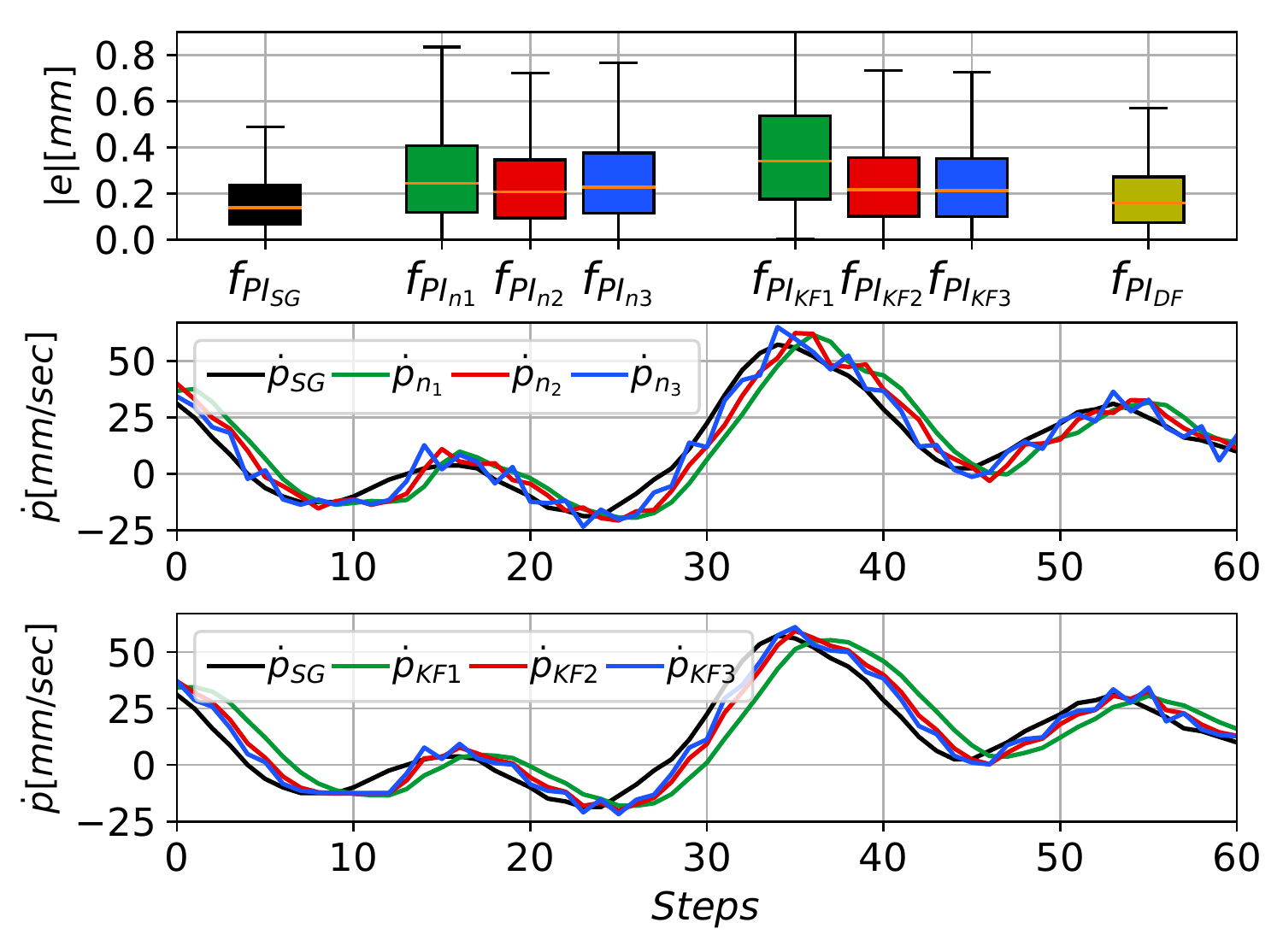}
    	\caption{\textcolor{black}{Comparison of the prediction errors obtained in the test set with physically inspired estimators, together with a detailed plot of the evolution of $\dot{p}$ computed by means of numerical differentiation and a Kalman filter.}}\label{fig:BB_enc_barplot}
    \begin{tabular}{|c|c|c|c|c|}
     \hline 
     & $f_{PI_{SG}}$ & $f_{PI_{n2}}$ & $f_{PI_{KF2}}$ & $f_{PI_{DF}}$\\
     \hline RMSE $[mm]$  & $0.2013$ & $0.2963$ & $0.3064$ & $0.2393$\\  \hline
   \end{tabular}
    \end{figure}
    
    \textcolor{black}{In Fig.~\ref{fig:BB_enc_barplot}, we visualize the distribution of the estimation errors module in the test set through boxplots, as well as reporting the numerical values of the $RMSE$. Acausal filtering guarantees the best performance, whereas, among the estimators with causal inputs, the proposed approach performs best. Indeed, the $RMSE$ obtained with the derivative-free estimator is approximately $20\%$ smaller than the best $RMSE$ obtained with the other causal estimators, i.e., $f_{PI_{n2}}$ and $f_{PI_{KF2}}$. As visible from the boxplots, the proposed solution exhibits a smaller variability. Results obtained with numerical differentiation and Kalman filtering show that the technique used to compute velocities can affect prediction performance significantly. In Fig.~\ref{fig:BB_enc_barplot}, we present also a detailed plot of the $\dot{p}$ evolution obtained with different differentiation techniques. As expected, there is a trade-off between noise rejection and delay introduced that must be considered. For instance, increasing the cutoff frequency decreases the delay, but at the same time impairs the rejection of noise. An inspection of the $f_{PI_{n1}}$, $f_{PI_{n2}}$ and $f_{PI_{n3}}$  prediction errors shows that too high or too low cutoff frequencies lead to the worst prediction performance. With our proposed approach, tuning is not required, since the filtering coefficients are learned automatically during the GPR training.}

\subsection{Ball-and-beam control}
\begin{figure}[t]
	\centering
	\includegraphics[width=\linewidth]{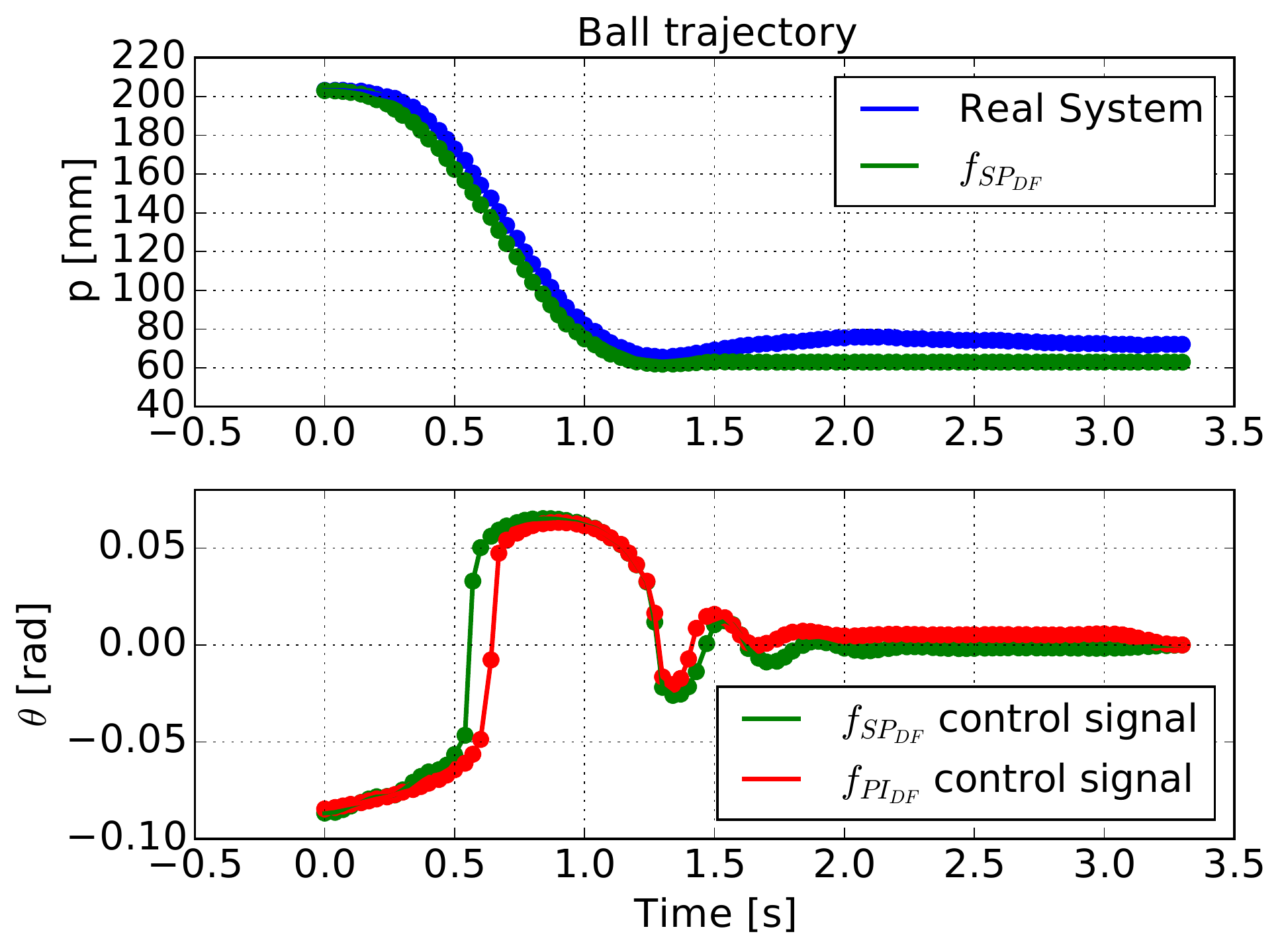}
	\caption{Top plot: comparison of the ball trajectory on the real system with the optimal trajectory computed by iLQG on $f_{SP_{DF}}$. Bottom plot: comparison of the control signals computed by iLQG using $f_{SP_{DF}}$ and~$f_{PI_{DF}}$.}
	\label{fig:BB_performance}
\end{figure}

The control task is the stabilization of the ball with zero velocity in a target position along the beam. The control trajectory is computed using the iLQG algorithm introduced in Section~\ref{subsec:iLQG}. In order to model also the behaviors not captured by the physical equations of motion, we train a GP, called $f_{SP_{DF}}$, with semiparametric kernel as in~Eq.\eqref{eq:SP_kernel_DF}:
\begin{equation}
\label{eq:SPDF_kernel_BB}
k^{BB}_{SPDF}(\boldsymbol{x_i}, \boldsymbol{x_j}) =  k^{BB}_{PIDF}(\boldsymbol{x_i}, \boldsymbol{x_j}) + k^{BB}_{NPDF}(\boldsymbol{x_i}, \boldsymbol{x_j}) \text{.}
\end{equation}
where the NP kernel is $k^{BB}_{NPDF}(\boldsymbol{x_i}, \boldsymbol{x_j}) = k_{RBF}(\boldsymbol{x_i}, \boldsymbol{x_j})$ with the $\Sigma_{RBF}$ matrix parameterized through Cholesky decomposition. The training data are the same described in Section~\ref{subsec:BB_pred_performance}. The control trajectory obtained by iLQG using $f_{SP_{DF}}$ model is applied to the physical system, and performance is shown in Fig.~\ref{fig:BB_performance}. 
In the top plot, we can observe how the optimized trajectory for the model remains close to the ball trajectory of the real system for all the 100 steps (3.3[s]), which is the chosen length for the iLQG trajectory. This result illustrates the high accuracy of the model in estimating the future evolution of the real system. Note that the control trajectory is implemented in open loop, to highlight the model precision obtaining an average deviation between the target and the final ball position of 9[mm] and standard deviation of 5[mm] in 10 runs. By adding a small proportional feedback control, the error becomes almost null. In the bottom plot, the control trajectory obtained by iLQG using either $f_{SP_{DF}}$ or $f_{PI_{DF}}$ is shown. Two major observations can be made: the trajectory obtained with $f_{SP_{DF}}$ approximates a bang-bang trajectory that in a linear system would be the optimal trajectory, and the trajectory obtained with $f_{PI_{DF}}$ is similar, but since the equation of motions cannot describe all the nonlinear effects present in a real system, the control action has a final bias that makes the ball drift away from the target position. 
\section{FURUTA PENDULUM: DERIVATIVE FREE MODELING AND CONTROL}
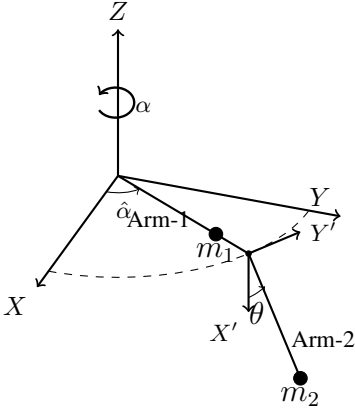
\begin{figure}[t]
	\centering
	\tdplotsetmaincoords{60}{110}
	
	\pgfmathsetmacro{\rvec}{.6}
	\pgfmathsetmacro{\thetavec}{90}
	\pgfmathsetmacro{\phivec}{60}
	\pgfmathsetmacro{\rvector}{.45}
	\pgfmathsetmacro{\tvector}{.25}
	
	\begin{tikzpicture}[scale=4.5,tdplot_main_coords]
	
	\coordinate (O) at (0,0,0);
	
	\tdplotsetcoord{P}{\rvec}{\thetavec}{\phivec}
	\tdplotsetcoord{P1}{\rvector}{\thetavec}{\phivec}
	\tdplotsetcoord{P2}{\tvector}{\thetavec}{\phivec}

	
	\draw[thick,->] (0,0,0) -- (0.7,0,0) node[anchor=north east]{$\large{X}$};
	\draw[thick,->] (0,0,0) -- (0,0.7,0) node[anchor=south east]{$\large{Y}$};
	\draw[thick,->] (0,0,0) -- (0,0,0.5) node[anchor=south]{$\large{Z}$};
	
	
	\draw[thick] (O) -- (P);
	\filldraw(P1) circle (0.02cm)node[align=left,below] {\large{$m_1$}};
	\filldraw(P) circle (0.008cm);
	\node[] at (0.2,0.2,0) {\small{Arm-1}};
	\draw (0,0,0)  -- (0,0,0.5)  node [midway] {\AxisRotator[y=0.2cm,z=0.4cm,->]};
	\node[] at (0,0.08,0.25) {\small{$\alpha$}};

	\tdplotdrawarc[->]{(O)}{0.1}{0}{\phivec}{anchor=north}{\small{$\hat{\alpha}$}}

	\tdplotsetthetaplanecoords{\phivec}
	
	
	\draw[dashed] (\rvec,0,0) arc (0:90:\rvec);
	
	\tdplotsetrotatedcoords{\phivec}{\thetavec}{0}
	
	\tdplotsetrotatedcoordsorigin{(P)}
	
	\draw[thick,tdplot_rotated_coords,->] (0,0,0) -- (.2,0,0) node[anchor=north east]{\small$X'$};
	\draw[thick,tdplot_rotated_coords,->] (0,0,0) -- (0,.2,0) node[anchor=west]{\small$Y'$};
	
	
	\draw[thick,tdplot_rotated_coords] (0,0,0) -- (.5,.2,0);
	\filldraw[tdplot_rotated_coords,draw=black,fill=black] (0.5,0.2,0) circle (0.02 cm)node[align=left,below] {\large{$m_2$}};
	\node[] at (0.7,0.9,0) {\small{Arm-2}};
	
	\tdplotdrawarc[tdplot_rotated_coords,->]{(0,0,0)}{0.15}{0}{25}{anchor=north,color=black}{\large$\theta$}
	
	\tdplotsetrotatedthetaplanecoords{45}
	
	
	\end{tikzpicture}
	\caption{
	A schematic diagram of the FP with various system parameters and state variables. Arm-$j$ with $j=\{1,2\}$ has length $L_j$, mass $m_j$, inertia $J_j$ and center of mass for the two arms at $l_j$.
	}\label{fig:scheme_furuta_pendulum}
	\vspace{-6mm}
\end{figure}

The second physical system considered is the Furuta pendulum \cite{furuta1992new}, a popular benchmark system in control theory
.
A schematic of the FP with its parameters and variables is shown in Fig.~\ref{fig:scheme_furuta_pendulum}. We refer to ``Arm-1'' and ``Arm-2'' in Fig.~\ref{fig:scheme_furuta_pendulum} as the base arm and the pendulum arm, respectively, and we denote $\hat{\alpha}$ and $\theta$ the angles of the base arm and the pendulum.

 In \cite{cazzolato2011dynamics}, the authors have presented a model of the FP. 
Based on that model, we obtained the expression of $\ddot{\theta}$ as a linear function w.r.t a vector of parameters $\boldsymbol{w}$,
\begin{small}
	\begin{align}
		\ddot{\theta} &= \frac{(-\ddot{\hat{\alpha}}m_2 L_1 l_2 \cos(\theta) + .5 \dot{\hat{\alpha}}^2\hat{J}_2 \sin(2\theta)+ b_2 \dot{\theta} + g m_2 l_2 \sin(\theta))}{\hat{J}_2} \notag\\
	&= \begin{bmatrix}
	-\ddot{\hat{\alpha}}m_2 L_1 l_2 \cos(\theta) & \dot{\hat{\alpha}}^2\sin(2\theta)& \dot{\theta}& \sin(\theta)
	\end{bmatrix} \boldsymbol{w} \notag \\	
	&=\boldsymbol{\phi_{\ddot{\theta}}}(\ddot{\hat{\alpha}}, \dot{\hat{\alpha}}, \dot{\theta}, \theta)^T\boldsymbol{w}  \text{,} \label{eq:linear_relation_acc}
	\end{align}
\end{small}%
where $\hat{J}_1= J_1 + m_1l_1^2 + m_2L_1^2$ and $\hat{J}_2= J_2 + m_2l_2^2$.

\begin{figure}[b]
	\centering
	\includegraphics[width=0.65\columnwidth]{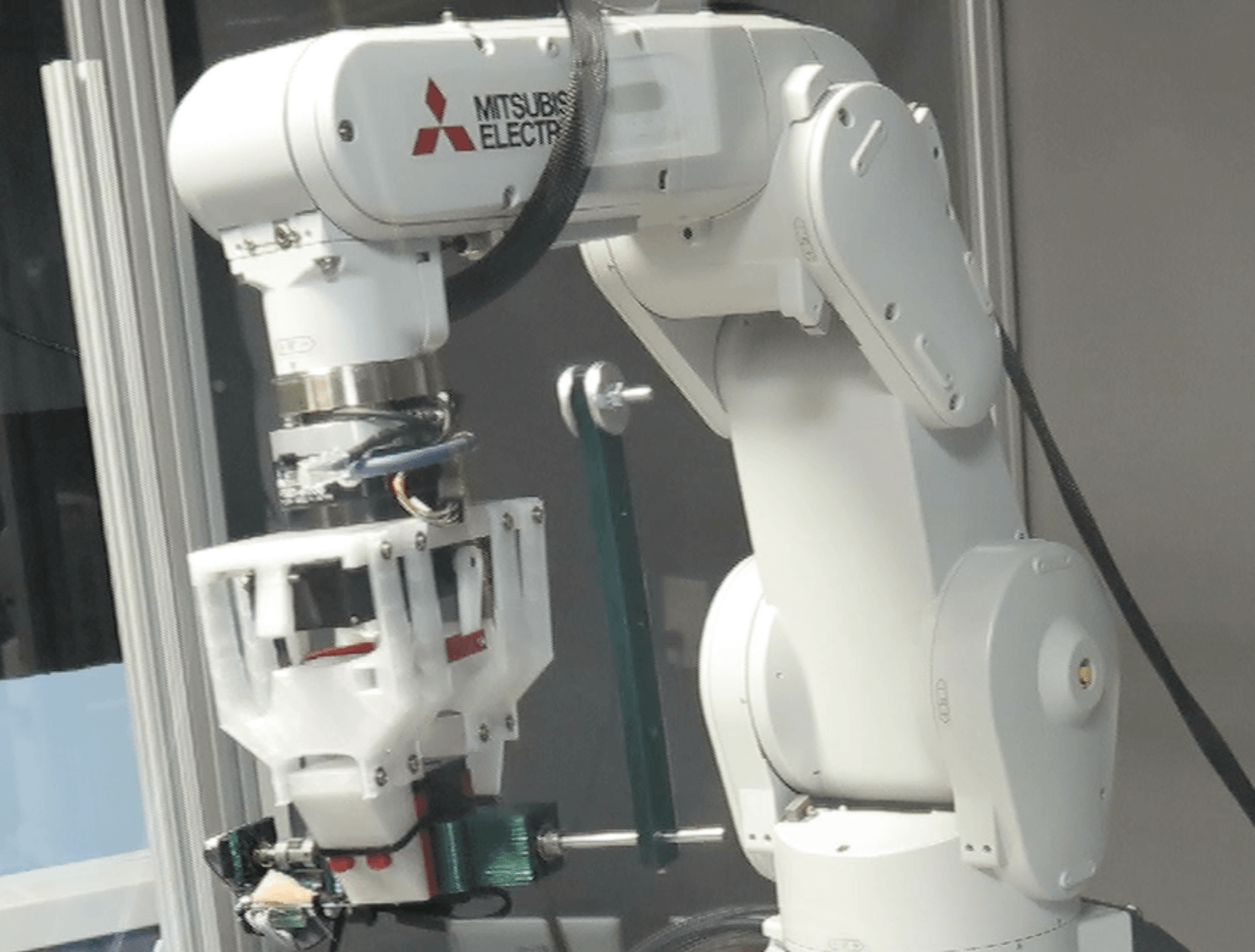}
	\caption{The Furuta Pendulum (3D printed in color green) held in the gripper at the swing-up position reached using the learned controller.}\label{fig:furuta_pendulum}
\end{figure}

The FP considered in this work has several characteristics that are different from those typically studied in the research literature. Indeed, in our FP (see Fig.~\ref{fig:furuta_pendulum}), the base arm is held by a gripper which is rotated by the wrist joint of a robotic arm (a MELFA RV-4FL). For this reason, the rotation applied to the wrist joint is denoted by $\alpha$, and it is different from the actual base arm angle $\hat{\alpha}$ (see Figure~\ref{fig:scheme_furuta_pendulum}). The control cycle of the robot is fixed at $7.1$ms, and communication to the robot and the pendulum encoder is handled by ROS.

These circumstances have several consequences. First, the robot can only be controlled in a position-control mode, and we need to design a trajectory of set points $\alpha^{des}$ considering that 
the manufacturer limits the maximum angle displacement of any robot's joint in a control period. 
This constraint, together with the high performance of the robot controller, results in a quasi-deterministic evolution of $\alpha$, that we identified to be 
$\alpha_k = (\alpha^{des}_{k}-\alpha^{des}_{k-1})/2$.
Therefore, the forward dynamics learning problem is restricted to model the pendulum arm dynamics. Additionally, the $3$D-printed gripper causes a significant interplay with the FP base link, due to the presence of elasticity and backlash. These facts lead to vibrations of the base arm along with the rotational motion, and a significant delay in actuation of the pendulum arm, which results in $\alpha\neq \hat{\alpha}$.

\subsection{Delay and nonlinear effects} \label{sec:exp_delay}
\begin{figure}[ht]
    \vspace{-2mm}
	\centering
	\includegraphics[width=.9\linewidth]{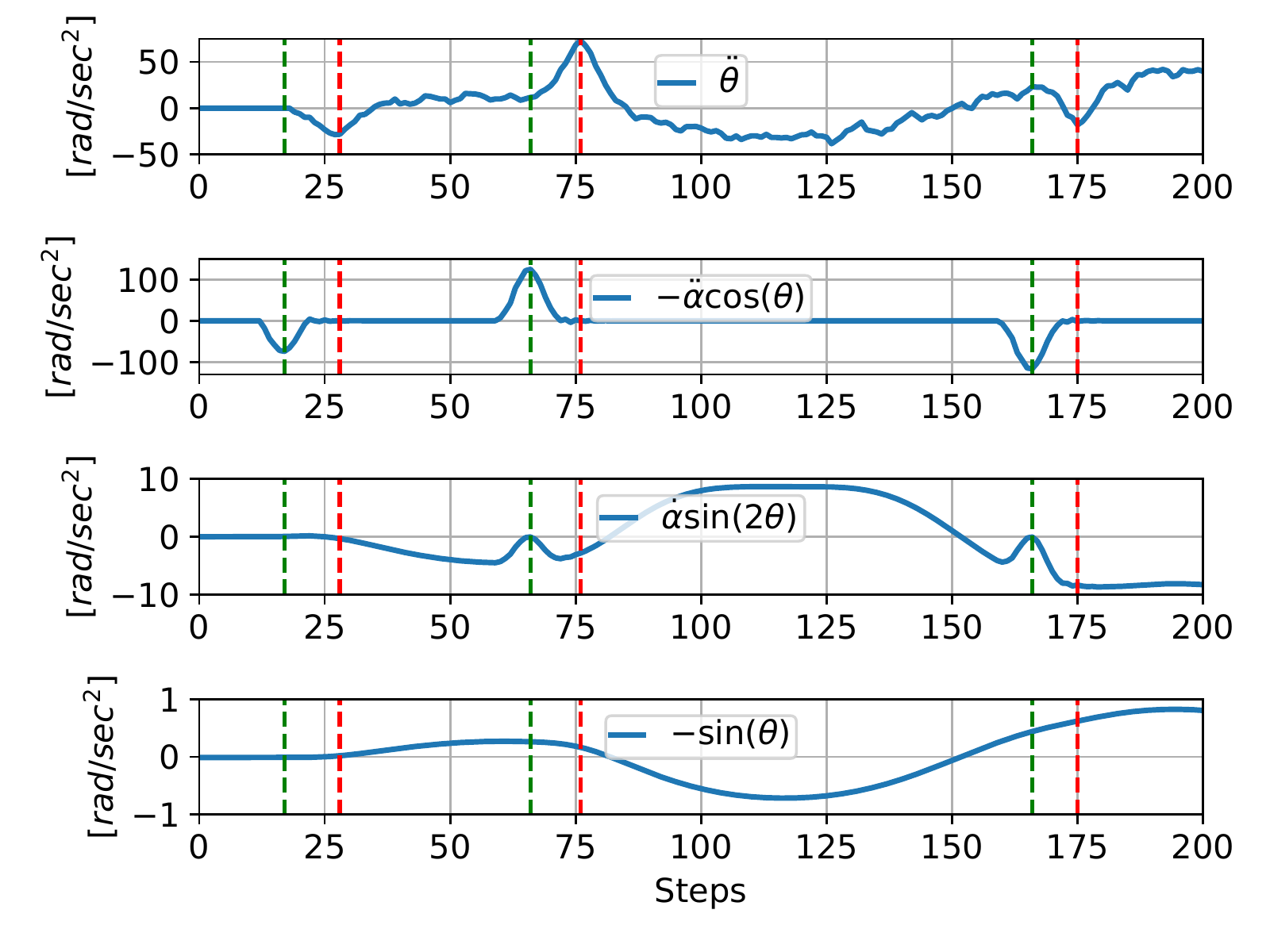}
	\caption{Evolution of $\ddot{\theta}$ and its model-based basis functions. Derivatives are computed using the acausal Savitzky-Golay filter.
	}
	\label{fig:delay}
	\vspace{-4mm}
\end{figure}

In order to demonstrate the presence of delays in the system dynamics, we report a simple experiment in which a triangular wave in $\alpha^{des}$ excites the system. The results are shown in Fig.~\ref{fig:delay}  
(for lack of space, the term depending on $\dot{\theta}$ is not reported, as the effects of viscous friction are not significant). 
The evolution of $\ddot{\theta}$ is characterized by a main low-frequency component with two evident peaks in the beginning of the trajectory, and a higher-frequency dynamical component which corrupts more the main component as the time passes by. Several insights can be obtained from these results. First, the peaks of the low-frequency component can be caused only by the $\ddot{\alpha}$ contribution, given that the $\dot{\alpha}$ and $\theta$ contributions do not exhibit these behaviours so prominently. Second, the difference between the peaks in the $\ddot{\alpha}$ contribution and $\ddot{\theta}$ (highlighted in the figure by the vertical dashed lines) represent the delay from the control signal and the effect on the pendulum arm. Third, the high-frequency component in $\ddot{\theta}$ might represent the noise generated by the vibration of the gripper, the elasticity of the base arm, and all the nonlinear effects given by the flexibility of the gripper.

\subsection{FP derivative free GPR models}\label{sec:FP_models}
We used the derivative-free framework to learn a model for the evolution of the pendulum arm. The FP state vector is defined as $\boldsymbol{x_k} = \left[\boldsymbol{x^{\theta}_k}, \boldsymbol{x^{\alpha}_k}, \alpha^{des}_{k-1}\right]$, with
\begin{equation*}
\boldsymbol{x^{\theta}_k} = \left[\theta_k,\dots,\theta_{k-k_p}\right] \text{,\,\,\,\,}
\boldsymbol{x^{\alpha}_k} = \left[\alpha_k,\dots,\alpha_{k-k_p}\right]\text{.}
\end{equation*}
From \eqref{eq:linear_relation_acc}, following the same procedure applied in the BB application to derive Eq.~\eqref{eq:linear_relation_delta_BB}, we obtain $\Delta_{\theta_k}=\theta_{k+1}-\theta_k = \boldsymbol{\phi_{\ddot{\theta}}}(\ddot{\hat{\alpha}}_k,\dot{\hat{\alpha}}_k,\dot{\theta}_k,\theta_k)^T\boldsymbol{w'}$.
Applying the guidelines in Section \ref{sec:guidelines} we obtain the corresponding PIDF kernel
\begin{small}
	\begin{align}
	k^{FP}_{PIDF}&(\boldsymbol{x_i},\boldsymbol{x_j}):= k_{poly}^1(\boldsymbol{x^{\alpha}_i},\boldsymbol{x^{\alpha}_j}) k_{poly}^1(\cos(\boldsymbol{x^{\theta}_i}),\cos(\boldsymbol{x^{\theta}_j})) \notag \\
	&+ k_{poly}^2(\boldsymbol{x^{\alpha}_i},\boldsymbol{x^{\alpha}_j})k_{poly}^1(\sin(2\boldsymbol{x^{\theta}_i}),\sin(2\boldsymbol{x^{\theta}_j})) \notag \\
	&+ k_{poly}^1(\sin(\boldsymbol{x^{\theta}_i}),\sin(\boldsymbol{x^{\theta}_j})) + k_{poly}^1(\boldsymbol{x^{\theta}_i},\boldsymbol{x^{\theta}_j}) \text{.} \label{eq:PI_kernel_FP}
	\end{align}
\end{small}%

In order to also model the complex behavior showed in Section~\ref{sec:exp_delay}, we define a semiparametric kernel for the FP~as:
\begin{equation}
\label{eq:SP_kernel_FP}
k^{FP}_{SPDF}(\boldsymbol{x_i}, \boldsymbol{x_j}) =  k^{FP}_{PIDF}(\boldsymbol{x_i}, \boldsymbol{x_j}) + k^{FP}_{NPDF}(\boldsymbol{x_i}, \boldsymbol{x_j}) \text{,}
\end{equation}
where the NP kernel is defined as the product of two RBFs with their $\Sigma_{RBF}$ matrices independently parameterized through Cholesky decomposition $k^{FP}_{NPDF}(\boldsymbol{x_i},\boldsymbol{x_j}) = k_{RBF}(\boldsymbol{x^{\alpha}_i},\boldsymbol{x^{\alpha}_j}) k_{RBF}(\boldsymbol{x^{\theta}_i},\boldsymbol{x^{\theta}_j})$.
Adopting a full covariance matrix, the RBF can learn convenient transformations of the inputs, increasing the generalization ability of the predictor. 

As experimentally verified in Section \ref{sec:exp_delay}, the evolution of $\theta$ is characterized by a significant delay w.r.t. the dynamics of $\alpha$. As a consequence, positions, velocities, and accelerations at time instant $k$ are not sufficient to describe the FP dynamics. However, defining the state as the collection of past measured positions, and setting properly $k_p$, the GPR has a sufficiently informative input vector, and can select inputs at the proper time instants, thus inferring the system delay from data. Note that when considering also velocities and accelerations, a similar approach would require a state of dimension $6k_p+1$, instead of $2k_p+1$.

\subsection{Prediction performance} \label{sec:exp_estimation}

In this section, we test the accuracy of different predictors:
\begin{itemize}[leftmargin=1.em]
	\item $f_{der}(\boldsymbol{x_k},\boldsymbol{\dot{x}_k},\boldsymbol{\ddot{x}_k})$: NP estimator defined in \eqref{eq:posterior_mean} with a RBF kernel with diagonal covariance and input given by $\boldsymbol{x_k}$ and its derivatives, i.e., all the positions velocities and accelerations from time $k$ to $k-k_p$, $k_p=15$; 
	\item $f_{NP}(\boldsymbol{x_k})$: NPDF estimator defined in \eqref{eq:posterior_mean} with a RBF kernel, $k_p=15$; 
	\item  $f_{PI}(\boldsymbol{x_k})$: the PIDF estimator defined in \eqref{eq:posterior_mean} with kernel defined in \eqref{eq:PI_kernel_FP}, $k_p=15$; 
	\item  $f_{SP}(\boldsymbol{x_k})$: the SPDF estimator defined in \eqref{eq:posterior_mean} with kernel defined in \eqref{eq:SP_kernel_FP}, $k_p=15$.
\end{itemize}
The $f_{der}$ model is considered to provide the performance of a standard NP estimator based on  $\boldsymbol{x^\alpha}$ and $\boldsymbol{x^\theta}$ derivatives.

\begin{figure}[ht]
	\centering
	\includegraphics[width=.9\linewidth]{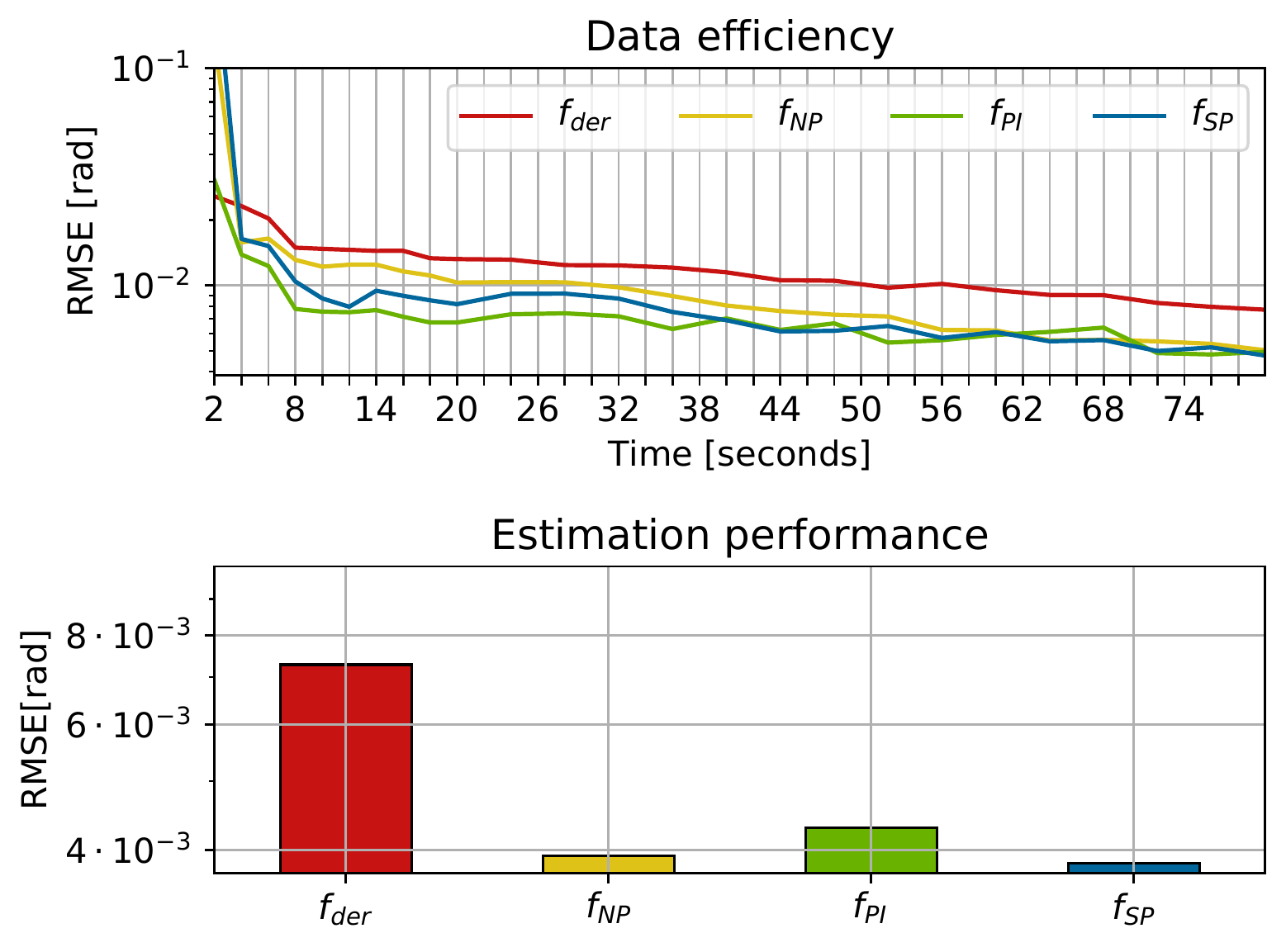}
	\caption{Top: evolution of the $RMSE$ as function of the experience seen by the model. Bottom: comparison of the $RMSE$ obtained training the estimators with all the data in $D_{tr}$. In both plots, the $y$-axis is in log-scale.
	}
	\label{fig:nRMSE}
	\vspace{-5mm}
\end{figure}

The estimators have been trained by minimizing the negative marginal log-likelihood (NMLL) over a training data set $D_{tr}$ composed of $15,000$ samples, corresponding approximately to $100$ seconds of experience. The input signal is a sum of $30$ sinusoids with random angular velocity ranging between $\pm8.5[rad/sec]$. To deal with the consistent number of samples available, we rely on stochastic gradient descent to optimize the NMLL. 
Performance is measured on a test data set $D_{sin}$ composed of $20,000$ samples, obtained with an input signal of the same type as the one considered in $D_{tr}$, but a different distribution of the sinusoids with frequency ranging between $\pm15[rad/sec]$, to show generalization ability. Estimators are compared both in terms of accuracy and data efficiency, and results are in Figure \ref{fig:nRMSE}. In the bottom graph, we report the Root Mean Squared Error ($RMSE$) in $D_{sin}$, and all the estimators considered are able to predict the evolution of the pendulum arm with an error smaller than one degree. However, derivative-free approaches outperform the non-derivative-free estimator. Note that $f_{SP}$ achieves the best performance, and $f_{NP}$ outperforms $f_{der}$, despite that both models are based on an RBF kernel. The latter fact confirms that numerical computation of the derivatives might reduce estimation accuracy. In the top graph, we report the evolution of the RMSE as a function of the seconds of the training samples available. The derivative-free approaches are more accurate and data-efficient than $f_{der}$. Notice that $f_{der}$ is more accurate only for the short period of the first $4$ seconds, and its RMSE decreases more slowly. The use of the PI kernel is particularly helpful as regards data efficiency, since after $2$ seconds of data $f_{PI}$ is more accurate than $f_{NP}$ and $f_{SP}$, and the $f_{SP}$'s $RMSE$ decreases faster than the one of $f_{NP}$.

\subsection{Rollout performance}
In this section, we characterize the rollout accuracy of the derived models, namely the estimation performance at $n$-step-ahead predictions. 
\begin{figure}[b]
	\centering
	\includegraphics[width=.88\linewidth]{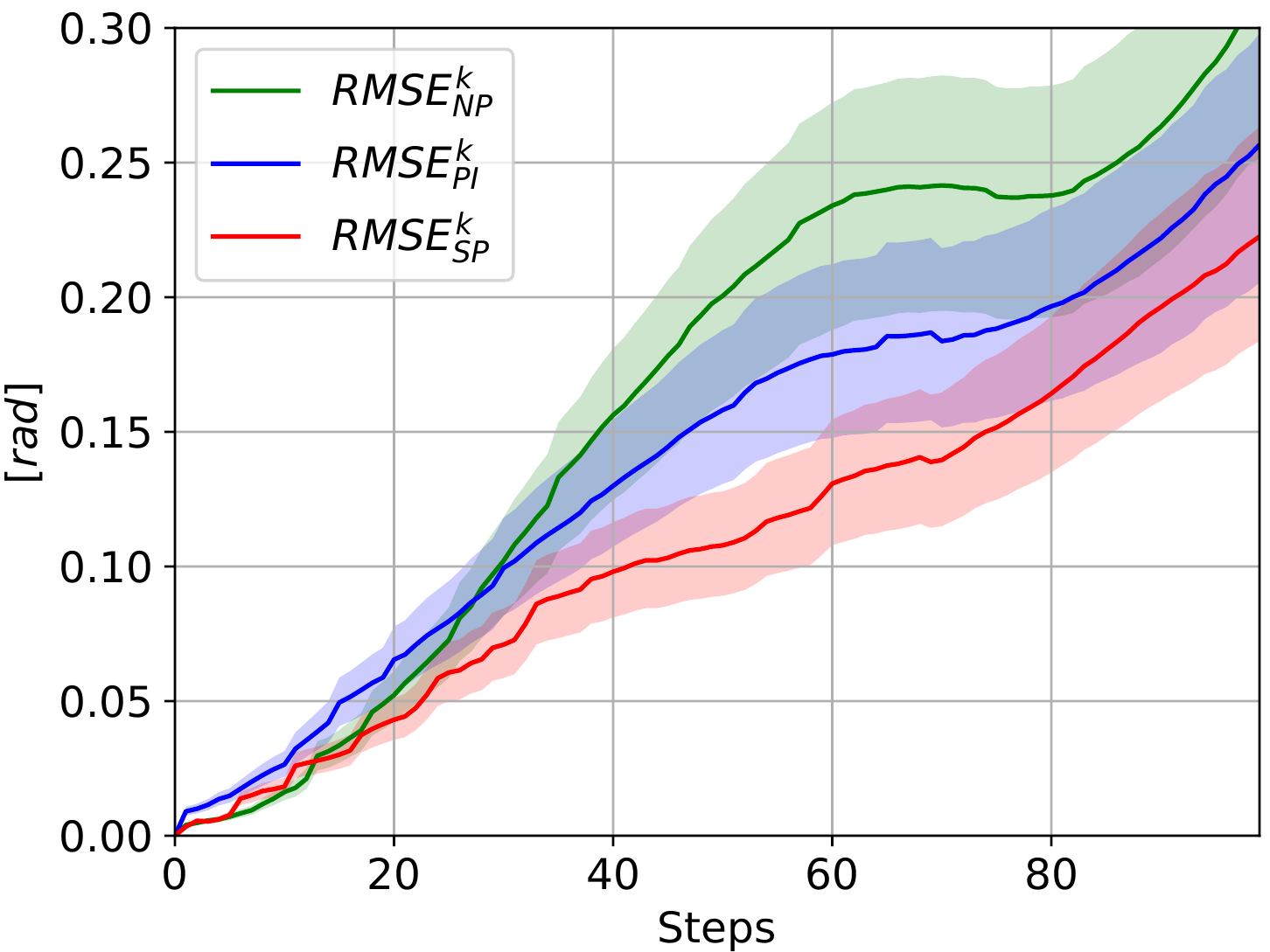}
	\caption{\textcolor{black}{Evolution of $RMSE^k$ with its $99\%$ confidence intervals.}}
	\label{fig:rolloutperf}
	\vspace{-4mm}
\end{figure}
For each model, we performed $N_{sim}$ rollouts. During the $i$-th rollout, we randomly pick an initial instant $k_i$, then the input location $\boldsymbol{x_{k_i}}$ in $D_{sin}$ is selected as initial input, and a prediction is performed for a window of $N_w=100$ steps. For each simulation, $\boldsymbol{e^i}=[e^i_1,\,\dots,\,e^i_{N_w}] \in \mathbb{R}^{N_w}$ is computed by subtracting the predicted trajectory from the one realized in $D_{sin}$. To characterize how uncertainty evolves over time, we define the error statistic $RMSE^k= \sqrt{\sum_{i=1}^{N_{sim}}\left(e^i_k\right)^2 \setminus N_{sim}}$,
that is the $RMSE$ of the prediction at the $k$-th step ahead. The $RMSE^k$ confidence intervals are computed assuming i.i.d. and normally distributed errors. Under this assumptions, each $RMSE^k$ has a $\chi^2$ distribution. 
The performance in terms of the $RMSE^k$ of  $f_{NP}$, $f_{PI}$ and $f_{SP}$ is reported in Fig.~\ref{fig:rolloutperf}.
In the initial phase, $RMSE^k_{f_{NP}}$ is lower than $RMSE^k_{f_{PI}}$, whereas for $k \simeq 30$ $RMSE^k_{f_{NP}}$ becomes greater than $RMSE^k_{f_{PI}}$. This suggests that the NP model behaves well for short interval prediction, whereas the PI model is more suitable for long-term predictions. 
The SP approach combines the advantages of these two models. The evolution of $RMSE^k_{f_{SP}}$ confirms this, showing that $f_{SP}$  outperforms $f_{NP}$ and $f_{PI}$.
\subsection{Furuta Pendulum control} \label{sec:control}
\begin{figure}[h]
   \vspace{-2mm}
	\centering
	\includegraphics[width=\linewidth]{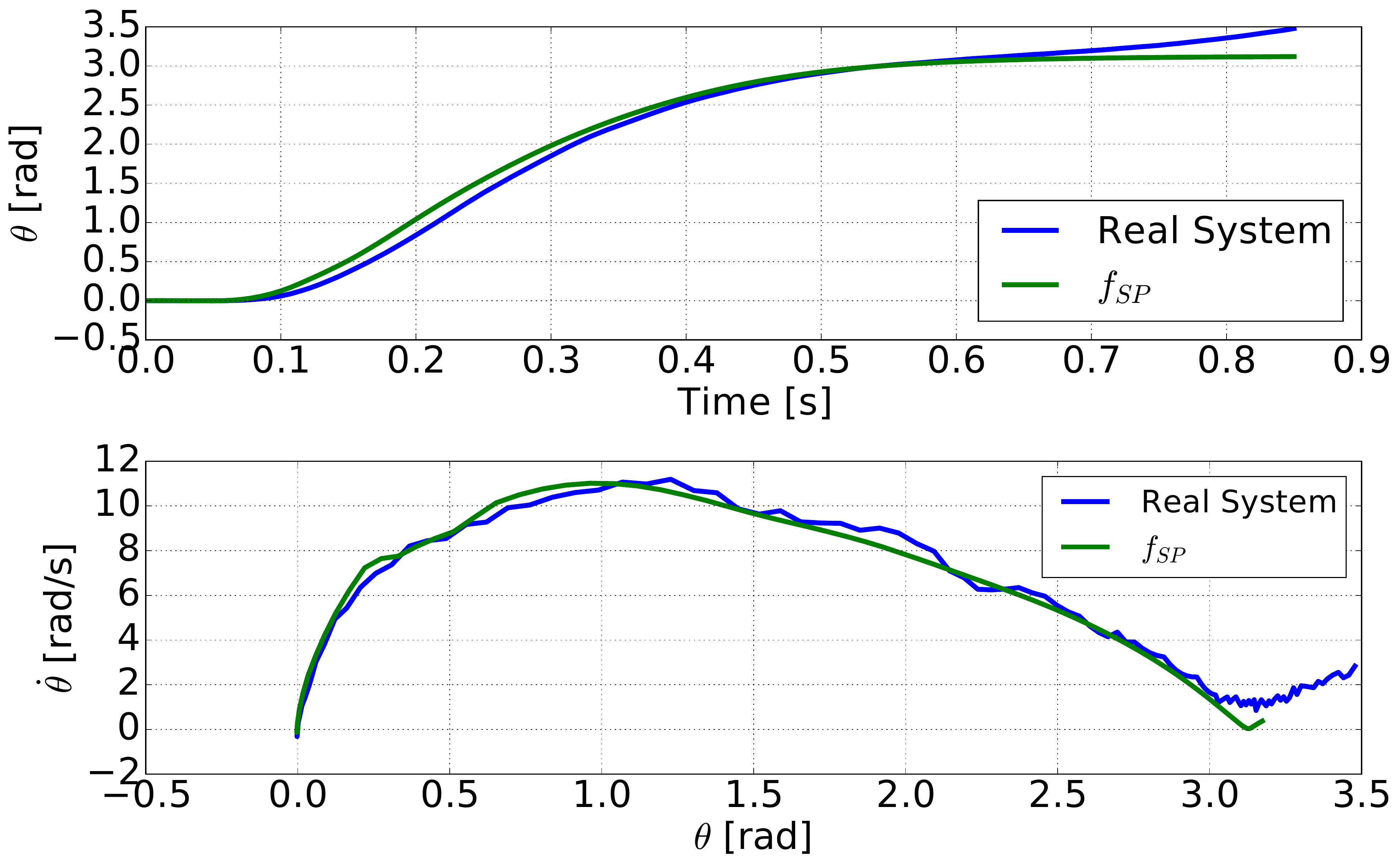}
	\caption{Performance of the iLQG trajectory on the FP swing-up control.
	}
	\label{fig:swing-up}
	\vspace{-4mm}
\end{figure}

The $f_{SP}$ model is used to design a controller to swing-up the FP using the iLQG algorithm described in Section~\ref{subsec:iLQG}.  
The model is accurate to the point that the trajectories obtained by the iLQG algorithm were implemented in an open-loop fashion on the real system, and the results are shown in Fig.~\ref{fig:swing-up}. The FP swings up with near-zero velocity~at the goal position; however, as expected, an open-loop control sequence cannot stabilize it. Fig.~\ref{fig:swing-up} reports the agreement between the $\theta$ trajectories obtained under the iLQG control sequence, using both the $f_{SP}$ and the real system. The comparison shows the long-horizon predictive accuracy of the learned model. Note that the models lose accuracy around the unstable equilibrium point, because of insufficient data, which are harder to collect in this area during training.
\section{CONCLUSIONS}
In this paper, we presented a derivative-free learning framework for model based RL, and we defined a novel physically-inspired derivative-free kernel. Experiments with two real robotic systems show that the proposed learning framework outperforms in prediction accuracy its corresponding derivative-based GPR model, and that semi-parametric derivative-free methods are accurate enough to solve model-based RL control problems in real-world applications.
The proposed framework exhibits robustness to delays and a capacity to deal with partially observable systems that can be further investigated.

\addtolength{\textheight}{-12cm}  




\bibliographystyle{IEEEtran}
\bibliography{references}

\end{document}